\documentclass[journal]{IEEEtran}
\usepackage{xcolor}
\usepackage{amsmath}
\usepackage{amssymb}
\usepackage{multirow}
\usepackage{booktabs}
\usepackage{makecell}
\newcommand{\ra}[1]{\renewcommand{\arraystretch}{#1}}
\usepackage{graphicx}
\usepackage{arydshln}
\usepackage[noadjust]{cite}

\ifCLASSINFOpdf
\else
\fi
\begin{document}
\title{Improving Face-Based Age Estimation with Attention-Based Dynamic Patch Fusion}

\author{Haoyi~Wang,
        Victor~Sanchez,~\IEEEmembership{Member,~IEEE,}
        Chang-Tsun~Li,~\IEEEmembership{Senior~Member,~IEEE}% <-this % stops a space
\thanks{H. Wang and V. Sanchez are with the Department
of Computer Science, University of Warwick, Coventry, CV4 7AL, UK (e-mail: h.wang.16@warwick.ac.uk, v.f.sanchez-silva@warwick.ac.uk.)}% <-this % stops a space
\thanks{C-T. Li is with the School of Information Technology, Deakin University, Geelong VIC 3216, Australia (e-mail: changtsun.li@deakin.edu.au.)}% <-this % stops a space
\thanks{© 2021 IEEE.  Personal use of this material is permitted.  Permission from IEEE must be obtained for all other uses, in any current or future media, including reprinting/republishing this material for advertising or promotional purposes, creating new collective works, for resale or redistribution to servers or lists, or reuse of any copyrighted component of this work in other works.}% <-this % stops a space
% \thanks{Manuscript received April 19, 2005; revised August 26, 2015.}}
}


\markboth{Journal of \LaTeX\ Class Files,~Vol.~14, No.~8, August~2015}%
{Shell \MakeLowercase{\textit{et al.}}: Bare Demo of IEEEtran.cls for IEEE Journals}



\maketitle

\begin{abstract}
With the increasing popularity of convolutional neural networks (CNNs), recent works on face-based age estimation employ these networks as the backbone. However, state-of-the-art CNN-based methods treat each facial region equally, thus entirely ignoring the importance of some facial patches that may contain rich age-specific information. In this paper, we propose a face-based age estimation framework, called Attention-based Dynamic Patch Fusion (ADPF). In ADPF, two separate CNNs are implemented, namely the AttentionNet and the FusionNet. The AttentionNet dynamically locates and ranks age-specific patches by employing a novel Ranking-guided Multi-Head Hybrid Attention (RMHHA) mechanism. The FusionNet uses the discovered patches along with the facial image to predict the age of the subject. Since the proposed RMHHA mechanism ranks the discovered patches based on their importance, the length of the learning path of each patch in the FusionNet is proportional to the amount of information it carries (the longer, the more important). ADPF also introduces a novel diversity loss to guide the training of the AttentionNet and reduce the overlap among patches so that the diverse and important patches are discovered.  Through extensive experiments, we show that our proposed framework outperforms state-of-the-art methods on several age estimation benchmark datasets.

\end{abstract}


% \fi
\IEEEpeerreviewmaketitle

\section{Introduction}

% outline
% 1. what the face-based age estimation is about
% 2. people start to use patch based methods, like ...
% 3. drawback of the above methods, i.e. they use fixed facial attributes
% 4. ICIP paper use age-specific patches
% 5. drawback of the ICIP paper. computational complexity of the Adaboost, location of patches are fixed for every image, and it cannot be trained in a end-to-end manner.
% 6. The new proposed method.
% 7. contributions

\IEEEPARstart{F}{ace}-based age estimation is an active and challenging research topic that keeps attracting attention from the research community \cite{geng2007automatic, guo2009human, chen2013subspace, li2014human, han2015demographic, hu2016facial, niu2016ordinal, han2017heterogeneous, wang2018fusion, angeloni2019age}. The aim of the face-based age estimation task is to predict the real age (accumulated years after birth) of a subject from their facial images. This task has several applications in diverse scenarios like security control, video surveillance, and merchandise recommendation \cite{fu2010age,wang2020using}.

Modern face-based age estimation methods typically consist of two components, a feature extractor and an estimator. The feature extractor is used to extract age-specific features from raw facial images and the estimator is used to predict the age based on the extracted features. Many recent works \cite{hu2016facial, feng2016human, niu2016ordinal, liu2017label, liu2017group, Chen_2017_CVPR, shen2018deep, pan2018mean, li2019bridgenet, cao2019consistent} focus on designing customized estimators while treating the facial image as an ordinary input, hence paying no attention to the relative importance of the extracted features. However, related studies \cite{guo2009human, han2015demographic, wang2018fusion} show that age-specific patches are useful when predicting the age of the subject from an image. In other words, customized feature extractors can be designed to exploit age-specific patches during training to boost the performance of face-based age estimation methods. As a consequence, many works now tackle the face-based age estimation problem by leveraging cropped age-specific patches as complementary inputs to their estimator \cite{yi2014age, han2015demographic, wang2018fusion, chen2019age, angeloni2019age}. The patches used in most of these works are those depicting dominant facial attributes like the eyes, nose, and mouth. However, early studies on face-based age estimation \cite{wu1995dynamic, boissieux2000simulation, akazaki2002age, mukaida2004extraction, lagarde2005topography, golovinskiy2006statistical, lai2013role} show that the most informative patches for this problem are where wrinkles typically appear, like eye bags and laugh lines. To locate these age-specific patches, Han \textit{et al.} \cite{han2015demographic} leverage the Bio-Inspired Features (BIF) proposed in \cite{guo2009human}. Later, Wang \textit{et al.} \cite{wang2018fusion} design a customized CNN to fuse the features learned from the facial image and the BIF-based patches. Unfortunately, the computed BIF-based patches in these methods are fixed in every image, which prevents extracting features that are robust to the location and shape variations of age-specific regions. 

% Yi \textit{et al.} \cite{yi2014age} first adopted cropped facial attributes in multiple scales. In their method, each cropped region is fed to one convolutional neural network (CNN) to extract features. \cite{chen2017multi} used a similar strategy in which each sub-network in the multi-path CNN takes a pair of images as the input. Each pair of image consists of the original facial image and a selected dominant attribute including the eyes, the nose and the mouth. Later, Chen \textit{et al.} extended this idea by using the original facial image alongside a dominant attribute to predict age, gender and race simultaneously.

% Although the above CNN-based methods leveraging dominant facial attributes in their methods,

% early studies on the age estimation problem \cite{wu1995dynamic, boissieux2000simulation, akazaki2002age, mukaida2004extraction, lagarde2005topography, golovinskiy2006statistical, lai2013role} show that the most informative regions for this particular task are where wrinkles typically appear, like eye bags and laugh lines. To locate these age-specific regions, \cite{han2015demographic} used Bio-Inspired Features (BIF) proposed in \cite{guo2009human}. However, since this method is not CNN-based, its performance is relatively poor compared to some aforementioned CNN-based methods.

\begin{figure}[t]
\begin{center}
\includegraphics[width=1\linewidth]{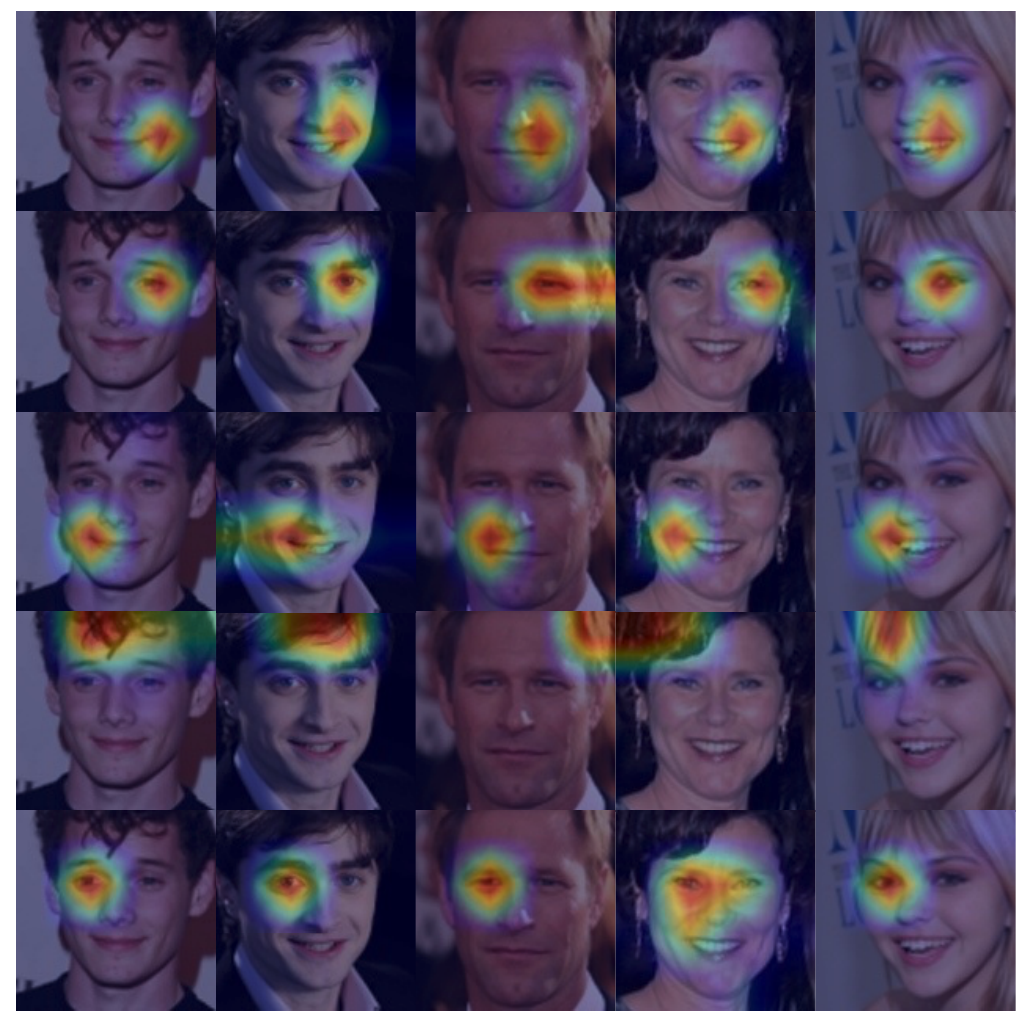}
\end{center}
   \caption{Five most informative age-specific patches, indicated as heat maps, used in ADPF. Each row depicts an age-specific patch from one head in RMHHA for five different samples. For each sample, patches are presented in a descending order in terms of importance from top to bottom.}
\label{fig:demo}
\end{figure}

In this paper, we propose a novel framework named Attention-based Dynamic Patch Fusion (ADPF) based on our preliminary work \cite{wang2018fusion} to tackle the face-based age estimation problem. ADPF comprises a customized feature extractor that consists of an AttentionNet and a FusionNet. The AttentionNet dynamically discovers age-specific patches by employing a novel attention mechanism, while the FusionNet predicts the age of the subject by fusing features learned from the facial image and the discovered age-specific patches. To improve performance, the discovered patches are fed into the FusionNet sequentially in a descending order based on the amount of age-specific information they carry. To this end, we introduce the Ranking-guided Multi-Head Hybrid Attention (RMHHA) mechanism into the AttentionNet. RMHHA is inspired by the Multi-Head Self-Attention (MHSA) mechanism \cite{vaswani2017attention}. However, instead of using the multi-channel feature maps produced by MHSA, each attention head in RMHHA yields a compact single-channel attention map, which is used to crop the corresponding age-specific patch from the facial image. RMHHA assigns a learnable weights to the produced attention maps to rank their importance. Hence, RMHHA not only helps to dynamically learn age-specific patches, but it also ensures the discovered patches are fed into the FusionNet in the desired order. The age-specific patches revealed by ADPF are exemplified in Fig. \ref{fig:demo}.

% In \cite{wang2018fusion}, due to the high-dimensional input to the patch acquisition algorithm, the process of computing and ranking age-specific patches is extremely time consuming. By replacing the patch acquisition algorithm in \cite{wang2018fusion} with the AttentionNet, the training time of the framework is significantly reduced. 

% In this work, we also introduce the Ranking-guided Multi-Head Hybrid Attention (RMHHA) mechanism, which is integrated into the AttentionNet. RMHHA learns dynamic age-specific patches and ensures the patches are fed into the FusionNet in the desired order. RMHHA is inspired by the Multi-Head Self-Attention (MHSA) mechanism \cite{vaswani2017attention}. However, instead of using the multi-channel feature maps produced by MHSA, each attention head in RMHHA yields a compact single-channel attention map, which is used to crop the corresponding age-specific patch from the facial image. To this end, we use hybrid attention in lieu of self-attention by combining the self-attention mechanism with the channel-wise attention mechanism. RMHHA assigns a learnable weight to the attention maps computed by the attention heads to rank their importance. The ranked patches are then fed into the FusionNet directly from the RMHHA. The age-specific patches revealed by our method are exemplified in Fig. \ref{fig:demo}.

The contributions of this paper are as follows:
\begin{itemize}
  \item We introduce ADPF, a framework that contains two networks, an AttentionNet and a FusionNet, to improve the face-based age estimation performance. 
%   Compared to our prior work \cite{wang2018fusion}, the ADPF has dramatically increased training efficiency.
  \item Instead of using the BIF and AdaBoost algorithms in \cite{wang2018fusion} to locate age-specific patches, ADPF uses the AttentionNet, which includes the novel RMHHA mechanism. RMHHA dynamically produces ranked single-channel attention maps, where each attention map highlights an age-specific patch.
  \item To reduce the overlap among patches, we propose a diversity loss to force RMHHA to reveal diverse age-specific regions.
  \item Through extensive experiments, we show that ADPF achieves state-of-the-art performance on several face-based age estimation benchmark datasets. We also show that, compared to our prior work \cite{wang2018fusion}, ADPF dramatically decreases training times.
\end{itemize}

The rest of this paper is organized as follows. In Section II, we review the related works on the face-based age estimation task and attention mechanisms. In Section III, we present the details of ADPF, including the RMHHA mechanism, the formulation of the diversity loss and the FusionNet. In Section IV, we explain the experimental settings and show the performance of ADPF and other state-of-the-art methods as evaluated on several age estimation benchmark datasets. Finally, we conclude our work in Section V.  

%% ------------------------------------------------------------------------------------------------- %%
%% ------------------------------------------------------------------------------------------------- %%
%% ------------------------------------------------------------------------------------------------- %%

\section{Related Work}

To lay the foundation of our work, in this section, related research on face-based age estimation is reviewed and discussed. We also review MHSA and other channel-wise attention mechanism, which are both related to the proposed RMHHA.

%% ------------------------------------------------------------------------------------------------- %%

\subsection{Face-based Age Estimation}

% Generally, existing methods for face-based age estimation consist of two components, a feature extractor and an estimator. The feature extractor aims at extracting compound high-dimensional age-specific features from facial images while the estimator is used to predict the age of the subject based on the extracted features. 

In the past few decades, many works have been conducted on face-based age estimation. One of the earliest works can be traced back to \cite{kwon1994age}, in which the researchers classify faces into three age groups based on the cranio-facial development theory and wrinkle analysis. Later, \cite{wu1995dynamic} reveals that wrinkles play an important role in modeling aging faces and determining ages.

Before deep learning-based methods dominated the computer vision field, researchers used to develop face-based age estimation methods with hand-crafted features. For example, the Statistical Face Model \cite{edwards1998statistical} used in \cite{lanitis2002toward} is adopted to extract features and reveal the relationship between features and the corresponding age labels. Geng \textit{et al.} \cite{geng2006learning, geng2007automatic} propose the AGing pattErn Subspace (AGES) to learn aging pattern vectors in a representative subspace from training images. Unseen faces are then projected to this newly constructed subspace to predict their ages. Later, \cite{gao2009face} reveals the ambiguity of mapping ages to age groups and proposes the Fuzzy Linear Discriminant Analysis (LDA) to build the classifier as an estimator. The authors define an Age Membership Function to encode the relevance between ages and age groups and integrate this function as a weighting factor into the conventional LDA. Guo \textit{et al.} \cite{guo2011simultaneous} propose a kernel-based regression method to tackle the face-based age estimation problem. A worth-noting algorithm designed to extract hand-crafted features for face-based age estimation is BIF \cite{guo2009human}. The BIF algorithm is based on the HMAX feature extraction method \cite{riesenhuber1999hierarchical}, which models the visual processing in the cortex. Specifically, it adopts the first two layers of HMAX, where the first layer convolves facial images with a set of Gabor filters \cite{gabor1946theory} and the second layer performs maximum (max) pooling over the features extracted by the first layer. The authors improve this bio-inspired method by adding a normalization operation after max pooling. They find that using only the first two layers of HMAX achieves better results in the age estimation scenario than using the entire HMAX method. Recently, Han \textit{et al.} \cite{han2015demographic} attach binary decision trees after the feature extraction process performed by the BIF algorithm to predict the age, gender and race simultaneously. 

% Only the first two layers from the prototype in~\cite{riesenhuber1999hierarchical} are used in this work, which achieve a better result for age estimation than using all the layers. After that, the dimension of the feature vector is reduced by using the Principle Component Analysis (PCA).

With the growing size of age-oriented datasets \cite{ricanek2006morph, chen2014cross}, CNNs are now the foundation of feature extraction methods. One of the first works to use CNNs for the face-based age estimation problem is \cite{wang2015deeply}, in which a CNN with two convolutional layers is deployed. Han \textit{et al.} \cite{han2017heterogeneous} use a modified AlexNet \cite{krizhevsky2012imagenet} to construct a multi-task learning method for heterogeneous face attribute estimation including the age. 

% Regardless the method researchers use, one of the design principles for age estimation methods is to leverage or design customized classification or regression methods to map the features with the corresponding age labels \cite{gao2009face, guo2009human, huang2012extreme}. To begin with, \cite{gao2009face} reveals the ambiguity of mapping ages to age groups and propose the Fuzzy Linear Discriminant Analysis (LDA) to build the classifier as an estimator. The authors define an Age Membership Function to encode the relevance between ages and age groups and integrate this function as a weighting factor into the conventional LDA. In the meantime, Guo \textit{et al.} \cite{guo2011simultaneous} propose a kernel-based regression method to tackle the face-based age estimation problem. 

In general, CNN-based face-based age estimation methods can be classified into two categories. Works in the first category aim to design customized estimators after the feature extraction stage to better model the mapping between the features and the corresponding age label. For example, Niu \textit{et al.} \cite{niu2016ordinal} treat face-based age estimation as an ordinal regression problem. In their work, a classifier with parallel fully-connected (FC) layers is constructed, where each FC layer produces a binary output that solves a binary classification sub-problem with respect to the corresponding age label. Chen \textit{et al.} \cite{Chen_2017_CVPR} also consider the ordinal relationship between different ages and propose the Ranking-CNN for face-based age estimation. Later, Pan \textit{et al.} \cite{pan2018mean} propose the mean-variance loss that consists of a mean loss and a variance loss aiming to learn a concentrated age distribution with a mean value close to the ground-truth. Recently, Shen \textit{et al.} \cite{shen2019deep} argue that the mapping between the facial features and the age label is inhomogeneous and introduce deep forests attached to CNNs to deal with such inhomogeneous mappings.

% leverage non-conventional classification or regression model as the  design a customized loss function to guide the training process.  

% Yi \textit{et al.} \cite{yi2014age} propose a multi-column CNN, which is one of the earliest works that apply CNNs to age estimation. 

While the aforementioned CNN-based methods focus on learning a sophisticated mapping between the features and the corresponding age label, the works in the second category try to boost performance with customized feature extractors. Yi \textit{et al.} \cite{yi2014age} propose a multi-stream CNN to better leverage high-dimensional structured information in facial images. The authors crop multiple patches from facial images so that each stream learns from one patch. Then, the features extracted from different patches are fused before the output layer. Angeloni \textit{et al.} \cite{angeloni2019age} and Chen \textit{et al.} \cite{chen2019age} also follow the same multi-stream CNN strategy. The patches used in these works are mainly dominant facial attributes such as the eyes, the nose, and the mouth, and not age-specific patches, which are those where wrinkles typically appear like eye corners and laugh lines \cite{wu1995dynamic, boissieux2000simulation, akazaki2002age, mukaida2004extraction, lagarde2005topography, golovinskiy2006statistical, lai2013role}.

To locate informative age-specific patches, our prior work \cite{wang2018fusion} uses the BIF and Adaboost algorithms. Specifically, the facial image is the primary input to the network as it carries more age-specific information than patches. The cropped patches are then subsequently fed into the CNN based on their importance, which is determined by the Adaboost algorithm. Due to the high-dimensional inputs to the BIF and Adaboost algorithms, the process of computing and ranking patches is extremely time consuming. In addition, the method proposed in our prior work consists of two separate stages, patch acquisition and CNN-training, which further increases the training complexity.

% The ADPF proposed in this paper is based on our prior work \cite{wang2018fusion} with several improvements. Inspired by works in natural language processing (NLP) field, we use the multi-head attention mechanism to discover multiple dynamic age-specific patches, which is much faster to compute compared to the patch acquisition process in \cite{wang2018fusion}. With the guide of the proposed RMHHA and the diversity loss, we directly fed age-specific patches into the FusionNet so that the whole method can be trained in an end-to-end manner with boosted performance. 

Since the facial image and cropped patches are processed by a different number of convolutional layers, i.e., the length of the learning path varies for different learning sources, the FusionNet in both our current and prior works involves fusing different levels of features. One work that also fuses different levels of features is \cite{xia2020multi}. However, the fused features in our work are from various inputs while the fused features in \cite{xia2020multi} are all from the input facial image.

\begin{figure*}
\begin{center}
\includegraphics[width=1\textwidth]{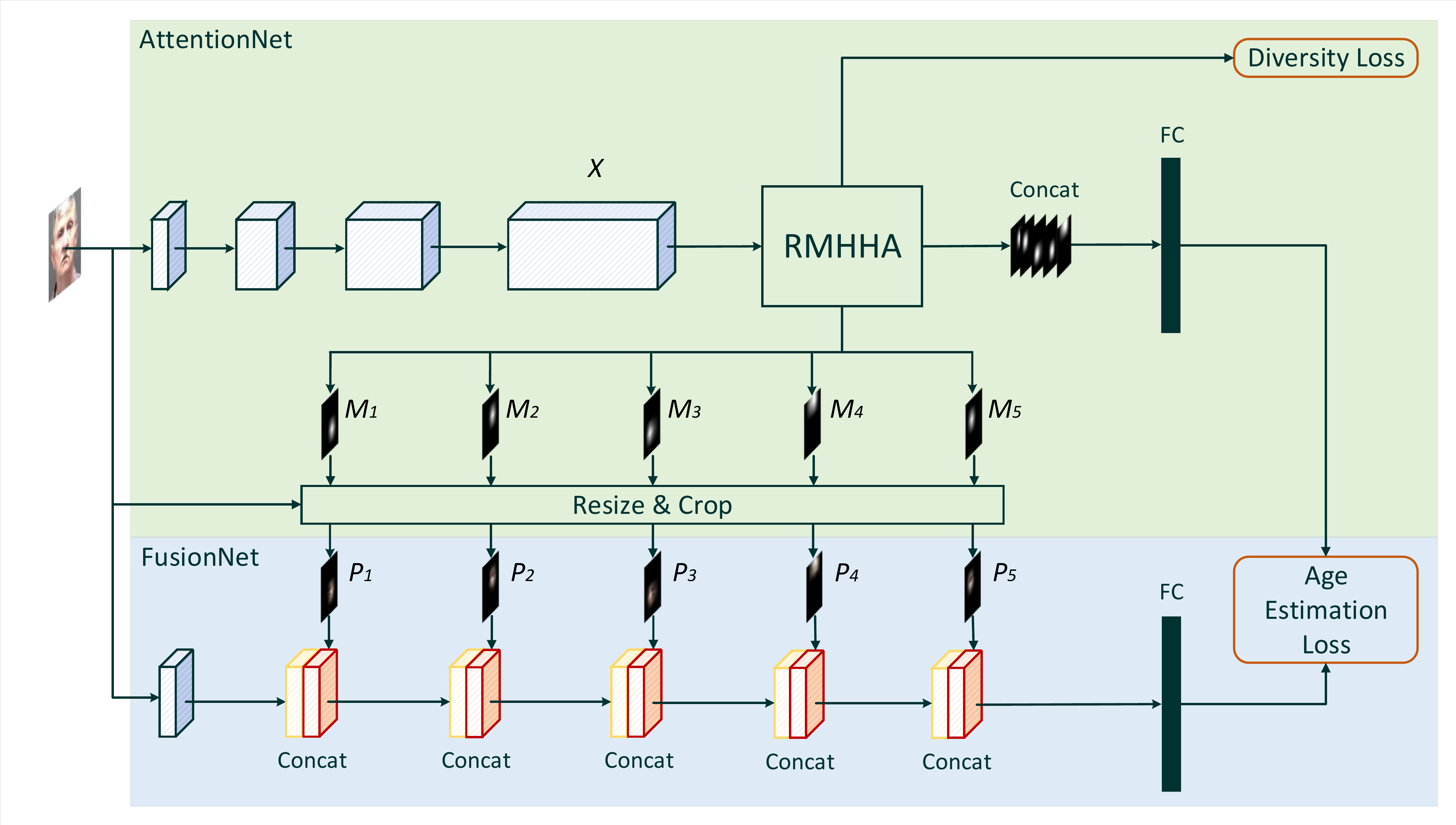}
\end{center}
   \caption{Architecture of ADPF. It consists of two networks, the AttentionNet and the FusionNet. The AttentionNet is used to train the proposed RMHHA to learn and rank age-specific features. Once the features are learned and ranked, denoted as $\boldsymbol{M1}$ to $\boldsymbol{M5}$ in the figure, we resize them to crop the corresponding patches from the input facial image. The cropped patches are listed as $\boldsymbol{\mathsf{P1}}$ to $\boldsymbol{\mathsf{P5}}$ in a descending order based on the amount of age-specific information they carry. Blocks represents CNN layers, $Concat$ indicates a concatenation operation, and $FC$ indicates a fully-connected layer. In particular, yellows blocks are from the previous layer in the main stream and red ones are from one particular age-specific patch. In addition, $\boldsymbol{\mathsf{X}}$ is the input tensor to the RMHHA mechanism with dimension of $32\times32\times500$.}
\label{fig:architecture}
\end{figure*}  

%% ------------------------------------------------------------------------------------------------- %%

\subsection{Attention Mechanisms}

\textbf{Multi-Head Self-Attention}. MHSA is first proposed in \cite{vaswani2017attention} and has been widely deployed as the backbone model for various Natural Language Processing (NLP) tasks \cite{devlin2019bert}. MHSA can attend to multiple informative segments of the input with an attention head attending to one specific segment. Therefore, the number of segments MHSA can attend to is determined by the number of attention heads.  MHSA has been recently used for imaging data. For example, Zhang \textit{et al.} \cite{zhang2019self} uses MHSA for the image synthesis task. Specifically, the authors propose the self-attention generative adversarial network (SAGAN) by adding MHSA layers to both the generator and the discriminator of a generative adversarial network (GAN) \cite{goodfellow2014generative}. With the help of MHSA layers, SAGAN can synthesize images with finer details than other state-of-the-art GAN models like \cite{brock2018large}. Several recent works \cite{bello2019attention,parmar2019stand} also use MHSA for image classification and object detection tasks.

\textbf{Channel-wise Attention}. Ever since Zeiler \textit{et al.} \cite{zeiler2014visualizing} visualized the feature maps learned by each channel in each layer of the AlexNet \cite{krizhevsky2012imagenet} trained on the ImageNet dataset \cite{deng2009imagenet}, researchers have been exploiting channel-wise attention mechanism to guide the network to pay attention to those channels that learn representative feature maps. Hu \textit{et al.} \cite{hu2018squeeze} integrate channel-wise attention into various CNN architectures \cite{simonyan2014very,szegedy2015going,he2016deep,howard2017mobilenets} to boost their performance on image classification and object detection tasks. Similarly, Zhang \textit{et al.} \cite{zhang2018image} and Chen \textit{et al.} \cite{chen2017sca} employ channel-wise attention to generate high-resolution images and image captions, respectively. Different from the aforementioned works where channel-wise attention is used to highlight informative channels in the input, in the proposed RMHHA mechanism, we use the computed channel-wise attention weights to merge the multi-channel self-attention maps into a single-channel attention map that reveals a particular age-specific patch.

Some previously proposed works also use attention mechanisms to discover multiple informative regions in an image. Specifically, Ba \textit{et al.} \cite{ba2015multiple} first use a recurrent neural network (RNN) to locate multiple regions, where each iteration of the recurrence outputs one region. Later, Chen \textit{et al.} \cite{chen2018recurrent}, Rao \textit{et al.} \cite{rao2017attention}, and Shi \textit{et al.} \cite{shi2019face} leverage reinforcement learning into the RNN to help the localization. However, such RNN-based methods produce regions with significant overlap, which may result in redundant post-processing. Different from these methods, we implement a diversity loss to force the attention mechanism in our model to reveal diverse regions.

% The non-local block in [9] is equivalent to the self-attention in transformers with an additional short-cut connection. The difference between the non-local block and the hybrid attention mechanism is that instead of using a short-cut connection, we use channel-wise attention to produce single-channel output.

Another related work is the one proposed in \cite{wang2017residual}, which uses a residual attention network to detect one or multiple objects in an image. However, their method cannot produce a consistent number of patches for each image, which is an important aspect of the attention mechanism in our model. 

It is worth noting that the hybrid attention mechanism proposed in this paper is similar to the non-local block in \cite{wang2018non}. The non-local block can be treated as the self-attention in transformers \cite{vaswani2017attention} with an additional short-cut connection. The difference between the non-local block and the hybrid attention mechanism is that instead of using a short-cut connection, we use channel-wise attention to produce a single-channel output.

% After a ranking operation, these discovered Patches are then fed into another network to produce the final output. In addition, we introduce a diversity loss to complement the training of the MHSA in order to avoid overlap among patches when the number of heads is large.

%% ------------------------------------------------------------------------------------------------- %%
%% ------------------------------------------------------------------------------------------------- %%
%% ------------------------------------------------------------------------------------------------- %%

\section{Attention-based Dynamic Patch Fusion}

In this section, we explain in detail ADPF by first discussing the core of the AttentionNet, i.e., the proposed RMHHA mechanism. Then, we formulate the diversity loss followed by explaining the FusionNet used to fuse features from various learning sources. The architecture of ADPF is illustrated in Fig. \ref{fig:architecture}.

\subsection{Ranking-guided Multi-Head Hybrid Attention}

% 1. Self-Attention
% 2. Hybrid Attention
% 3. Multi-head Hybrid Attention
% 4. Ranking-guided Multi-head Hybrid attention

Since RMHHA is based on MHSA and the key component in MHSA is the self-attention mechanism, we first discuss the self-attention mechanism followed by the proposed hybrid attention mechanism. Then, we detail the complete RMHHA mechanism.

Let us consider an input tensor $\boldsymbol{\mathsf{X}}$ that has a dimension of ${h}\times{w}\times{c}$, where $h$ denotes the height, $w$ denotes the width and the $c$ denotes the number of channels. $\boldsymbol{\mathsf{X}}$ is convolved into three separate tensors: $\boldsymbol{\mathsf{Q}}$ with a shape of ${h}\times{w}\times{c_{\boldsymbol{\mathsf{Q}}}}$, $\boldsymbol{\mathsf{K}}$ with a shape of ${h}\times{w}\times{c_{\boldsymbol{\mathsf{K}}}}$, and $\boldsymbol{\mathsf{V}}$ with a shape of ${h}\times{w}\times{c_{\boldsymbol{\mathsf{V}}}}$, where $c_{\boldsymbol{\mathsf{Q}}}$, $c_{\boldsymbol{\mathsf{K}}}$, and $c_{\boldsymbol{\mathsf{V}}}$ indicate the number of channels in the corresponding tensor. The intuition behind self-attention is to compute a weighted summation of the values, $\boldsymbol{\mathsf{V}}$, where the weights are computed as the similarities between the query, $\boldsymbol{\mathsf{Q}}$, and the corresponding key, $\boldsymbol{\mathsf{K}}$. Therefore, in order to compute the similarity, $\boldsymbol{\mathsf{Q}}$ and $\boldsymbol{\mathsf{K}}$ normally have the same shape, i.e., $c_{\boldsymbol{\mathsf{Q}}}=c_{\boldsymbol{\mathsf{K}}}$. The output of a single self-attention mechanism is computed as:
\begin{equation}\label{self_attention}
\begin{aligned}
    \boldsymbol{\mathsf{SA}}=Softmax(\frac{\boldsymbol{\mathsf{Q}}{'}\cdot\boldsymbol{\mathsf{K}}{'}^T}{\sqrt{c_{\boldsymbol{\mathsf{K}}}}})\cdot\boldsymbol{\mathsf{V}},
\end{aligned}
\end{equation}
where $\boldsymbol{\mathsf{Q}}'$ and $\boldsymbol{\mathsf{K}}'$ are flattened tensors in order to perform the dot product.

% To perform the self-attention on high-dimensional tensors, we flatten the input tensor into $X^{flatten}$ with a shape of ${HW}\times{C}$.

% For multi-head attention with $N$ heads, $X_{flatten}$ is equally divided into $N$ chunks. The input tensor for each head is denoted by $X_{head}$ with a dimension of ${H}\times{W}\times{C_{head}}$, where ${C_{head}=C/N}$. 

After the scaling operation, i.e., dividing the similarity matrix $\boldsymbol{\mathsf{Q}}' \cdot \boldsymbol{\mathsf{K}}'^T$ by a factor of $\sqrt{c_{\boldsymbol{\mathsf{K}}}}$ and applying the softmax function, we perform a dot product between the normalized similarity matrix and $\boldsymbol{\mathsf{V}}$ to generate the self-attention maps $\boldsymbol{\mathsf{SA}}$ with a dimension of ${h}\times{w}\times{c_{\boldsymbol{\mathsf{K}}}}$.

% where formally $Q=X^{flatten}W^{Q}_{n}$, $K_{n}=X^{'}_{n}W^{K}_{n}$, and $V_{n}=X^{'}_{n}W^{V}_{n}$. $W^{Q}_{n}$, $W^{K}_{n}$ and $W^{V}_{n}$ are parameter matrices of linear transformations where $W^{Q}_{n}\in\mathbb{R}^{C_{head}\times{C_{Q}}}$, $W^{K}_{n}\in\mathbb{R}^{C_{head}\times{C_{K}}}$, and $W^{V}_{n}\in\mathbb{R}^{C_{head}\times{C_{V}}}$. 

Since we flatten two-dimensional feature maps into an one-dimensional vector in Eq. \ref{self_attention}, the original structure of the feature maps is therefore distorted. To make it efficient when dealing with structured data like images and multi-dimensional features, we adopt the relative positional encoding in \cite{shaw2018self} and \cite{bello2019attention}. Specifically, the relative positional encoding is represented by the attention logit, which encodes how much an entry in $\boldsymbol{\mathsf{Q}}'$ attends to an entry in $\boldsymbol{\mathsf{K}}'$. The attention logit is computed as:
\begin{equation}\label{logit}
\begin{aligned}
    \boldsymbol{l}_{i,j}=\frac{\boldsymbol{q}_i^T}{\sqrt{c_{\boldsymbol{\mathsf{K}}}}}(\boldsymbol{k}_j+\boldsymbol{r}_{j_x-i_x}^w+\boldsymbol{r}_{j_y-i_y}^h),
\end{aligned}
\end{equation}
where $\boldsymbol{q}_i$ is the $i$-th row in $\boldsymbol{\mathsf{Q}}'$ indicating the feature vector for pixel $i:=(i_x,i_y)$ and $\boldsymbol{k}_j$ is the $j$-th row in $\boldsymbol{\mathsf{K}}'$ indicating the feature vector for pixel $j:=(j_x,j_y)$. $\boldsymbol{r}_{j_x-i_x}^w$ and $\boldsymbol{r}_{j_y-i_y}^h$ are learnable parameters encoding the positional information within the relative width $j_{x}-i_{x}$ and relative height $j_y-i_y$. With the relative positional encoding, the output of a single self-attention mechanism can be reformulated as:
\begin{equation}\label{sa_logit}
\begin{aligned}
    \boldsymbol{\mathsf{SA}}=Softmax(\frac{\boldsymbol{\mathsf{Q}}' \cdot \boldsymbol{\mathsf{K}}'^T+\boldsymbol{m}_h+\boldsymbol{m}_w}{\sqrt{c_{\boldsymbol{\mathsf{K}}}}}) \cdot \boldsymbol{\mathsf{V}},
\end{aligned}
\end{equation}
where $\boldsymbol{m}_h[i,j]=\boldsymbol{q}_i^T\boldsymbol{r}_{j_y-i_y}^h$ and $\boldsymbol{m}_w[i,j]=\boldsymbol{q}_i^T\boldsymbol{r}_{j_x-i_x}^w$ are matrices of relative positional logits.

\begin{figure*}
\begin{center}
\includegraphics[width=1\textwidth]{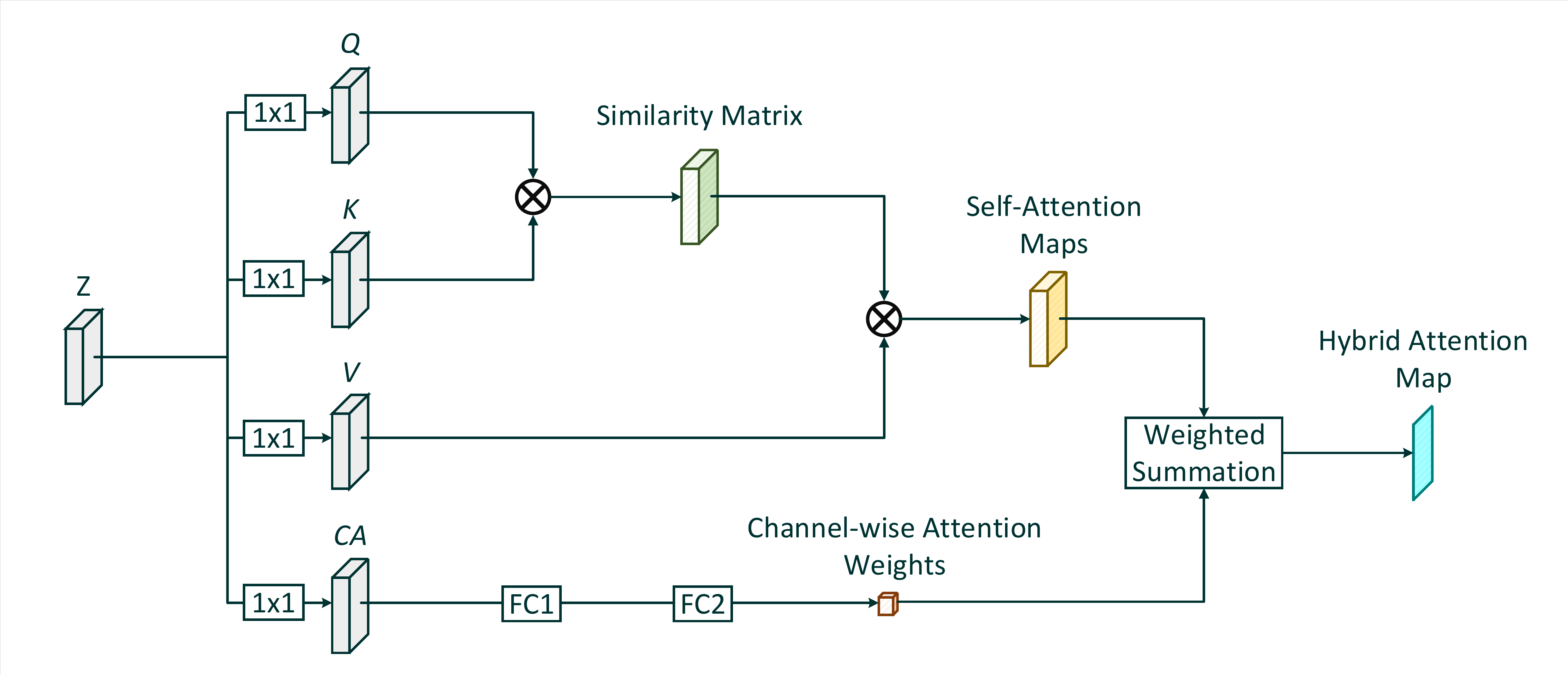}
\end{center}
   \caption{Structure of the proposed hybrid attention mechanism. $\boldsymbol{\mathsf{Q}}$, $\boldsymbol{\mathsf{K}}$, and $\boldsymbol{\mathsf{V}}$ are the \textit{query}, $keys$, and $value$, respectively, for the self-attention mechanism, and $CA$ is the input tensor to the channel-wise attention mechanism. The final hybrid attention map is computed as weighted summation, where the input tensor comprises the attention maps from the self-attention mechanism and the weights are computed from the channel-wise attention mechanism. 1x1 represents convolutional layers with kernel size of 1 and FC1 and FC2 indicate two fully-connected layers.}
\label{fig:hybrid_attention}
\end{figure*}

The output of the self-attention mechanism in Eq. \ref{sa_logit} has a dimension of $h\times{w}\times{c_{\boldsymbol{\mathsf{V}}}}$. However, we want each attention head to produce a single-channel attention map to depict one particular age-specific patch. To this end, we use channel-wise attention alongside self-attention to form a hybrid attention mechanism. Channel-wise attention is used to compute weights for each channel and a weighted summation is performed along the channel axis of the self-attention maps to generate the final single-channel attention map, indicated as the hybrid attention map in Fig. \ref{fig:hybrid_attention}.

As depicted in Fig. \ref{fig:hybrid_attention}, in the proposed hybrid attention mechanism, we first use a 1x1 convolutional layer on the input tensor, $\boldsymbol{\mathsf{Z}}$, to ensure the number of channels before computing the channel-wise attention weights matches the number of channels in the self-attention maps, i.e., $c_{\boldsymbol{\mathsf{V}}}$. The tensor after this 1x1 convolution is denoted as $\boldsymbol{\mathsf{CA}}$. We then aggregate each feature map in $\boldsymbol{\mathsf{CA}}$ with a pooling operation to produce a feature vector, in which each entry represents the features for the corresponding channel. Different from \cite{hu2018squeeze,Chen_2017_CVPR}, in which average pooling is used, we use max pooling to emphasize the most important features with high activation values. Following the procedure in \cite{hu2018squeeze}, we use a gating mechanism with two sequential FC layers to form a bottleneck. The first FC layer reduces the dimentionality, i.e., the number of channels, and the second FC layer increases the dimentionality of the previous layer to match the original shape. The output from the second FC layer is the set of channel-wise attention weights that we need, which are computed as:
\begin{equation}\label{ca}
\begin{aligned}
    \boldsymbol{w}_{CA} = \sigma(\boldsymbol{W}_{FC2}\delta(\boldsymbol{W}_{FC1}\delta(\boldsymbol{\mathsf{CA}}))),
\end{aligned}
\end{equation}
where $\delta$ indicates the non-linear ReLU function, $\sigma$ refers to the Sigmoid function used to normalize the attention weights, and $\boldsymbol{W}_{FC1}$ and $\boldsymbol{W}_{FC2}$ are learnable parameters in the two FC layers.

After the self-attention maps and channel-wise attention weights are computed, we perform a weighted summation over these two tensors along the channel dimension to get the single-channel hybrid attention map. The hybrid attention map is then computed as:
\begin{equation}\label{hybrid_attention}
\begin{aligned}
    \boldsymbol{HA} = \sum_{c}^{c_{\boldsymbol{\mathsf{V}}}}\boldsymbol{\mathsf{SA}}_{c}\cdot\boldsymbol{w}_{CA_c},
\end{aligned}
\end{equation}
where $c$ is the channel index and $\boldsymbol{\mathsf{SA}}$ is computed using Eq. \ref{sa_logit}.

To perform hybrid attention in a multi-head manner, each hybrid attention head takes a certain number of feature maps from the previous convolutional layer as the input. Specifically, assume there are $c_{p}$ feature maps in the tensor produced by the previous layer. Then, we have $c_{p}=c_{head}\times{n}$, where $n$ denotes the number of heads.

\begin{figure}[t]
\begin{center}
\includegraphics[width=1.\linewidth]{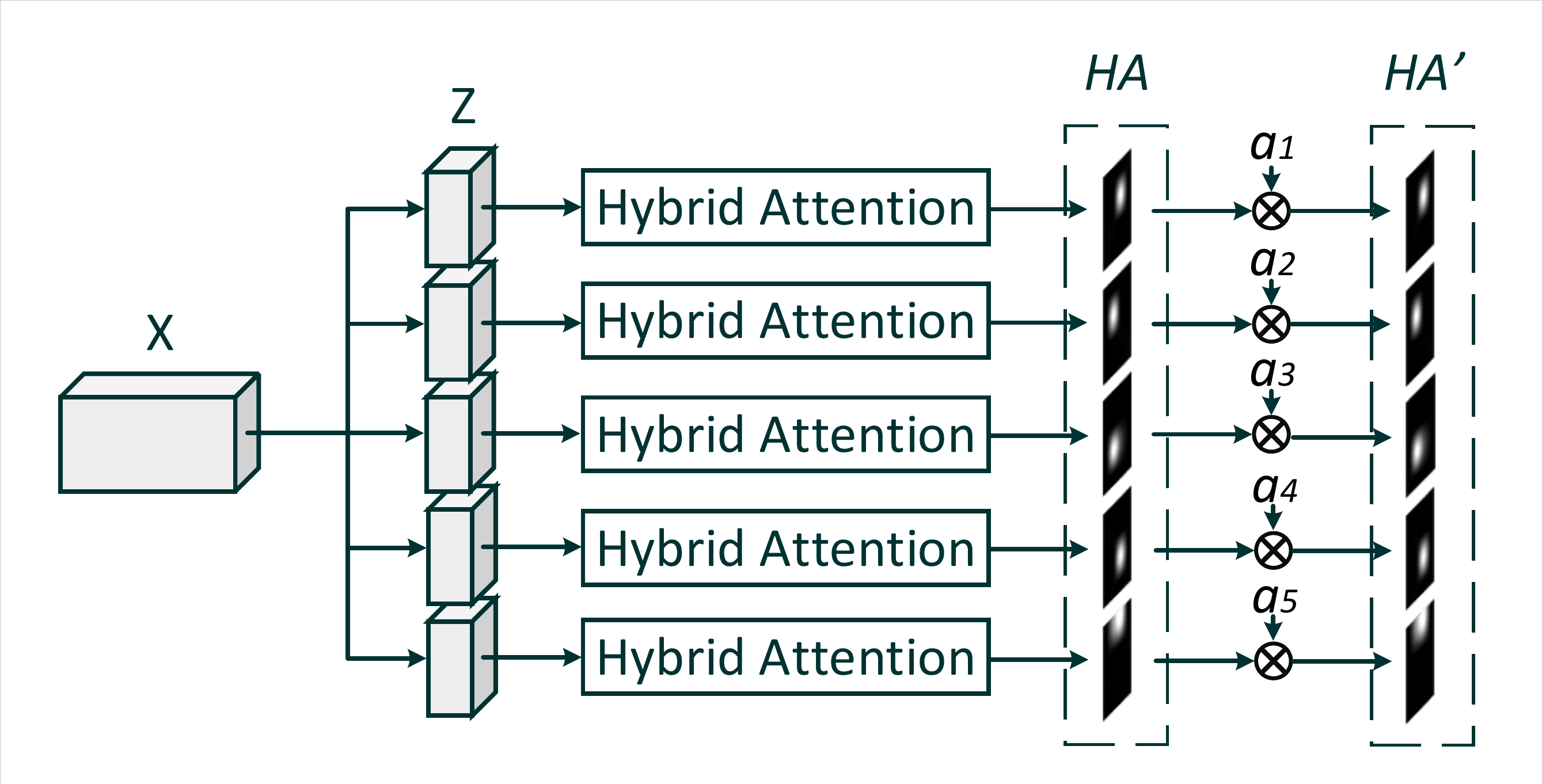}
\end{center}
   \caption{Architecture of the proposed RMHHA, where five attention heads are implemented.}
\label{fig:rmhha}
\end{figure}

Different from MHSA \cite{vaswani2017attention}, in which the attention maps from each head are concatenated right after the attention operation, we assign a learnable scale to each hybrid attention map to rank their importance when predicting ages, as shown in Fig. \ref{fig:rmhha}. RMHHA can then be formulated as:
\begin{equation}\label{rmhha}
\begin{aligned}
    \boldsymbol{\mathsf{RMHHA}} = \{\boldsymbol{HA}_{1}\cdot{a_{1}}, \boldsymbol{HA}_{2}\cdot{a_{2}}, ..., \boldsymbol{HA}_{n}\cdot{a_{n}}\},
\end{aligned}
\end{equation}
where $a_{n}$ indicates the learnable scale, which is updated by using the age estimation loss function presented in subsection III.D. $\boldsymbol{HA}_{n}\cdot{a_{n}}$ is equivalent to $\boldsymbol{HA}_n'$ in Fig. \ref{fig:rmhha}. All weighted hybrid attention maps used in ADPF are then concatenated before the final FC layer in the AttentionNet.

It is worth noting that multi-head attention methods always involve heavy matrix multiplications, which may be computationally expensive especially when the input matrices have a high dimentionality, which is common in CNNs. Therefore, differently from \cite{bello2019attention,parmar2019stand}, which stack dozens of MHSA models to compute the output, our work only uses one multi-head attention model to discover age-specific patches. 

% Moreover, in order to produce single-channel attention map, in our proposed RMHHA, we replace the self-attention with a novel hybrid attention by combining the self-attention with the channel-wise attention mechanism.

\subsection{Diversity Loss}

The number of patches that can be discovered is determined by the number of attention heads implemented in RMHHA. However, during implementation, we find that when using more than four heads, patches tend to overlap especially in informative regions. As demonstrated in Section IV, without further supervision, two attention maps may overlap in the nose region. This overlap of attended patches may led to redundant learning sources and leave other age-specific patches undiscovered. To alleviate this overlap issue, we propose a diversity loss to learn diverse and non-overlapping patches by minimizing the summation of products of corresponding entries in two hybrid attention maps, $\boldsymbol{HA}_{n_1}$ and $\boldsymbol{HA}_{n_2}$. The diversity loss is formulated as:
\begin{equation}\label{diversity}
\begin{aligned}
     \mathcal{L}_{diversity} = \sum_{\substack{n_1,n_2 \\ n_1\ne{n_2}}}^{n}\sum_{h'}^{h}\sum_{w'}^{w}\boldsymbol{HA}_{n_1}(h',w')\cdot\boldsymbol{HA}_{n_2}(h',w'),
\end{aligned}
\end{equation}
where $(h',w')$ denotes the location of the corresponding entry in a hybrid attention map.

\subsection{FusionNet}
The architecture of the FusionNet is illustrated in Fig. \ref{fig:architecture}. Since the attention maps (e.g., $\boldsymbol{\mathsf{M_1}}$ to $\boldsymbol{\mathsf{M_5}}$) produced by the AttentionNet only contain the location of the patches, they do not contain sufficient information to aid in the learning process of the FusionNet. It is then necessary to retrieve the corresponding patches from the input image. To get these patches, i.e., $\boldsymbol{\mathsf{P_1}}$ to $\boldsymbol{\mathsf{P_5}}$, we first rank the learned hybrid attention maps based on their associated weights, i.e., $a_1$ to $a_5$. $\boldsymbol{M_1}$ has the highest weight indicating that the corresponding age-specific patch represents the most age-specific information. After the hybrid attention maps are ranked, they are resized into the same spatial size as the original facial image and used to crop the corresponding highlighted area by performing the contour detection based on the boundary in attention maps.

Instead of training separate shallow CNNs for each input and concatenating the information before the final FC layer, we merge the features in the convolution stage. In the FusionNet, the length of the path to learn from an input is directly proportional to the amount of information it carries. This approach also allows extracting and emphasizing common age-specific features among all inputs. For example, the skin feature, which has an ordinal relationship with the age, can be emphasized since all inputs are expected to share the same skin texture. 

% The adoption of concatenation is inspired by the DenseNets \cite{huang2017densely}. In a DenseNet, the network is divided into several dense blocks, and layers within the same block typically share the identical spatial dimension. More importantly, inside each dense block, the output of each layer flows directly into all of the subsequent layers. As a result, the \(l\)-th layer receives feature maps from all the previous layers within the same block as the input:
% \begin{equation}\label{densenet}
%     x_{l}=D_{l}([x_{0}, x_{1},...,x_{l-1}])
% \end{equation}
% where $x$ represents the output of each layer, $D_{l}$ denotes the learning hypothesis of the $l$-th layer, and the $concat[\cdot]$ indicates the concatenation operation.

In the FusionNet, we preform concatenation operations on pairs of feature maps, one from the previous layer in the main stream (yellow blocks in Fig. \ref{fig:architecture}), $\boldsymbol{\mathsf{I}}$ , and the other representing the features learned from one particular age-specific patch (red blocks in Fig. \ref{fig:architecture}), $\boldsymbol{\mathsf{P}}$. Therefore, the concatenation in the FusionNet is formulated as:
\begin{equation}\label{block}
    \boldsymbol{\mathsf{R}} = Concate[\boldsymbol{\mathsf{I}}, \boldsymbol{\mathsf{P}}].
\end{equation}
This formulation is also commonly used in modern CNN architectures like the ResNet \cite{he2016deep} and the DenseNet \cite{huang2017densely}. Therefore, a sub-stream in the FusionNet can be treated as a shortcut connection, which emphasizes the learning of the age-specific information shared by all inputs.

\subsection{Age Estimation Loss}

To estimate the age, we use a regression loss to learn the exact age and a divergence loss to learn the age distribution (i.e., the label distribution learning \cite{shen2017label}). Specifically, after the features are processed by a Softmax function, we eliminate all the negative values in the output vector and normalize the remaining values so that they can form a probability distribution that sums up to 1:
\begin{equation}\label{norm}
     o_{p}:=
    \begin{cases}
	0 & o_{t}\le{0} \\
	\frac{\sum_{p=1}^{q}\max(0,o_{p})}{o_{p}} & o_{t}>0,
	\end{cases} 
\end{equation}
where $o_{p}$ is the $p$-th element in the output vector $\boldsymbol{o}\in{\mathbb{R}^q}$ and $q$ is the total number of classes.

The final prediction is the summation of products of the probabilities by the corresponding age labels:
\begin{equation}\label{estimation}
    E = \sum_{p=1}^{q}o_{p}g_{p},
\end{equation}
where $o_{p}$ denotes the normalized probability from Eq. \ref{norm} and $g_{p}$ is the associated age label for class $p$.

We use the mean absolute error (MAE) to compute the error between the prediction and the corresponding ground truth label:
\begin{equation}\label{mae}
    \mathcal{L}_{MAE} = \frac{1}{b}\sum_{b'}^{b}|E_{b'}-GT_{b'}|,
\end{equation}
where $b$ is the batch size and $GT$ refers to the ground truth label.

Recent works \cite{hu2016facial, pan2018mean} also include a soft label technique to model the age distributions. Specifically, since there is no noticeable visual change of a face over a few years, adopting such technique can explicitly create more training samples for each label (e.g. age). Following these works, we use the KL-divergence to measure the difference between a Gaussian distribution derived from the label \cite{pan2018mean} and the learned distribution. The KL-divergence is formulated as:
\begin{equation}\label{kl}
    \mathcal{L}_{KL} = \sum_{p=1}^{q}P(p)log\left(\frac{P(p)}{P'(p)}\right),
\end{equation}
where $P$ is the ground truth distribution and $P'$ is the learned distribution. The complete age estimation loss is then defined as a summation of these two losses:
\begin{equation}\label{ae}
    \mathcal{L}_{AE} = \mathcal{L}_{MAE} + \mathcal{L}_{KL}.
\end{equation}

\subsection{Training Strategy}

Since the training of the FusionNet requires well-learned and stabilized patches, we first train the AttentionNet with RMHHA until convergence. The overall loss to train this network is the summation of two loss functions:
\begin{equation}\label{attentionnet}
    \mathcal{L}_{AttentionNet} = \mathcal{L}_{AE} + \lambda\mathcal{L}_{diversity},
\end{equation}
where $\lambda$ controls the relative importance between two learning objectives.

When the AttentionNet converges, we freeze its parameters and start training the FusionNet. The loss function used to train the FusionNet is the loss formulated in Eq. \ref{ae}.

%% ------------------------------------------------------------------------------------------------- %%
%% ------------------------------------------------------------------------------------------------- %%
%% ------------------------------------------------------------------------------------------------- %%

\section{Experiments}

\subsection{Dataset}

\begin{table}
\ra{1.05}
\begin{center}
\label{tab:datasets}
\caption{Statistics of Three Benchmark Datasets}
\begin{tabular}{l@{\hskip 0.3in}c@{\hskip 0.3in}c@{\hskip 0.3in}c}\toprule
\hfil{Dataset} & \hfil{\#images} & \hfil{\#subjects} & \hfil{age range}  \\ \midrule
\hfil{MORPH II} & \hfil{55,134} & \hfil{13,618} & \hfil{16-77} \\
\hfil{FG-NET} & \hfil{1,002} & \hfil{82} & \hfil{0-69} \\
\hfil{CACD} & \hfil{163,446} & \hfil{2000} & \hfil{16-62} \\
\bottomrule
\end{tabular}
\end{center}
\end{table}

We conduct experiments on three commonly used face-based age estimation benchmark datasets, the MORPH II dataset \cite{ricanek2006morph}, the FG-NET dataset \cite{cootes2008fg}, and the Cross-Age Celebrities Dataset (CACD) \cite{chen2014cross}. 

The MORPH II dataset contains more than 55,000 facial images from about 13,000 subjects with ages ranging from 16 to 77 and an average age of 33. The distribution of race labels in the MORPH II dataset is extremely unbalanced as more than 96\% of subjects are annotated as \textit{African} or \textit{European} and individuals from \textit{Asia} and other regions only occupy less than 4\%. Each facial image in the MORPH II dataset is associated with identity, age, race and gender labels. 

The FG-NET dataset has 1002 facial images belonging to 82 subjects. Each subject in this dataset has more than 10 facial images taken over a long time span. In addition, the facial images in this dataset contain pose, illumination and expression (PIE) variations. 

The CACD contains more than 160,000 facial images from 2000 celebrities with ages ranging from 16 to 62. Similar to the images in the FG-NET dataset, facial images in the CACD contain PIE variations. The characteristics of these three datasets are presented in Table I.

\subsection{Experimental Settings}

\textbf{Data Pre-processing}. We use the open-source computer vision library dlib~\cite{dlib09} for image pre-processing. Firstly, 68 facial points are detected in each facial image to crop them based on the location of the eyes to a size of \(128\times{128}\) pixels.

Further, data augmentation is used to increase the dataset size. Specifically, images are zero-padded first and then cropped to the original size. Finally, the cropped images are randomly flipped horizontally.

\begin{table}
\ra{1.20}
\begin{center}
\caption{MAE values for several state-of-the-art Face-based Age Estimation Methods on the MORPH II Dataset under Setting I. Params indicates the number of parameters.}
\label{tab:morph1}
\begin{tabular}{p{0.18\textwidth}p{0.11\textwidth}p{0.11\textwidth}}\toprule
\hfil{Method} & \hfil{Params} & \hfil{MAE}\\ \midrule
\hfil{OHRank \cite{chang2011ordinal}} & \hfil{-} & \hfil{6.07} \\
\hfil{IIS-LLD \cite{geng2013facial}} & \hfil{-} & \hfil{5.67} \\
\hfil{CPNN \cite{geng2013facial}} & \hfil{-} & \hfil{4.87} \\
\hfil{OR-SVM \cite{chang2010ranking}} & \hfil{-} & \hfil{4.21} \\
\hfil{BFGS-LDL \cite{geng2016label}} & \hfil{-} & \hfil{3.94} \\
\hfil{OR-CNN \cite{niu2016ordinal}} & \hfil{~ 7M} & \hfil{3.27} \\
\hfil{DEX \cite{rothe2018deep}} & \hfil{~ 138M} & \hfil{3.25} \\
\hfil{SMMR \cite{huang2017soft}} & \hfil{-} & \hfil{3.24} \\
\hfil{ARN \cite{agustsson2017anchored}} & \hfil{~ 139M} & \hfil{3.00} \\
\hfil{Ranking-CNN \cite{Chen_2017_CVPR}} & \hfil{~ 26M} & \hfil{2.96} \\
\hfil{MSFCL \cite{xia2020multi}} & \hfil{~ 15M} & \hfil{2.90} \\
\hfil{DAG-GoogleNet \cite{taheri2019use}} & \hfil{~ 24M} & \hfil{2.87} \\
\hfil{DAG-VGG16 \cite{taheri2019use}} & \hfil{~ 131M} & \hfil{2.81} \\
\hfil{Mean-Variance Loss \cite{pan2018mean}} & \hfil{~ 20M} & \hfil{2.80} \\
\hfil{MSFCL-LR \cite{xia2020multi}} & \hfil{~ 15M} & \hfil{2.79} \\
\hfil{Hu \textit{et al.} \cite{hu2016facial}} & \hfil{~ 24M} & \hfil{2.78} \\
\hfil{BIF + FusionNet \cite{wang2018fusion}} & \hfil{~ 5M} & \hfil{2.76} \\
\hfil{MSFCL-KL \cite{xia2020multi}} & \hfil{~ 15M} & \hfil{2.73} \\
\hfil{VDAL \cite{liu2020similarity}} & \hfil{~ 19M} & \hfil{2.57} \\
\hfil{ADPF (ours)} & \hfil{{~ 14M}} & \hfil{\textbf{2.54}} \\
\bottomrule
\end{tabular}
\end{center}
\end{table}

\textbf{Dataset Partition}. For the MORPH II dataset, three commonly used settings are adopted. In the first setting, i.e., \textit{Setting I}, following prior works \cite{xia2020multi,taheri2019use,wang2018fusion,pan2018mean,Chen_2017_CVPR,liu2017label,niu2016ordinal}, we randomly split the whole dataset into two subsets, one with 80\% of the data for training and the other with 20\% for testing. In this setting, there is no identity overlap between the two subsets. To perform statistical analysis, we use 20 different partitions (with the same ratio but different distribution) and report mean values. In the second setting, i.e., the \textit{Setting II}, to compensate for the imbalance of race distribution, we randomly split the dataset into three subsets, denoted as \textit{S1}, \textit{S2}, and \textit{S3}, and ensure the ratio between Black and White labels is 1:1 and that between Male and Female labels is 1:3. In order to follow the same protocol as other works \cite{li2019bridgenet,chen2019age,chen2017multi,yi2014age,guo2011simultaneous}, the results under this setting are reported in three different ways: 1) training on \textit{S1} and testing on \textit{S2+S3}; 2) training on \textit{S2} and testing on \textit{S1+S3} and 3) the average value from the previous two scenarios. Finally, in the third setting, i.e., the \textit{Setting III}, we select 5,492 facial images of White people to reduce the variance caused by imbalanced race distribution \cite{rothe2018deep,agustsson2017anchored,guo2008image,wang2015deeply}. Then, these 5,492 facial images are randomly split into two subsets, 80\% of the them are used for training and the remaining 20\% for testing. To further reduce the data distribution variance, in this setting, we use 5-fold cross validation to produce the final results.

For the FG-NET dataset, we use the leave-one-person-out (LOPO) strategy \cite{xia2020multi,shen2019deep,liu2017label,lu2015cost,geng2013facial,geng2006learning}. In each fold, we use facial images of one subject for testing and the remaining images for training. Since there are 82 subjects, this process consists of 82 folds and the reported results are the average values.

For the CACD, following the setup in \cite{chen2017multi,rothe2018deep,shen2019deep}, the whole dataset is divided into three subsets, denoted as the training set, validation set, and testing set. The training set has facial images from 1,800 subjects, the validation set has facial images from 120 subjects, and the testing set has facial images from 80 subjects. The reported results are computed by training either on the training set or the validation set and evaluating on the testing set.

\begin{table}
\ra{1.20}
\begin{center}
\caption{MAE values for several state-of-the-art Face-based Age Estimation Methods on the MORPH II Dataset under Setting II.}
\label{tab:morph2}
\begin{tabular}{cp{0.08\textwidth}p{0.08\textwidth}p{0.08\textwidth}}\toprule
\multirow{2}{*}{Method} & \multicolumn{3}{c}{MAE} \\
\cmidrule{2-4}
&\hfil{S1/S2+S3} &\hfil{S2/S1+S3} &\hfil{Average} \\ \midrule
\hfil{KPLS \cite{guo2011simultaneous}} & \hfil{4.21} & \hfil{4.15} & \hfil{4.18} \\
\hfil{MS-CNN \cite{yi2014age}} & \hfil{3.63} & \hfil{3.63} & \hfil{3.63} \\
\hfil{MRNPE (AlexNet) \cite{chen2017multi}} & \hfil{2.98} & \hfil{2.73} & \hfil{2.86} \\
\hfil{MRNPE (VGG16) \cite{chen2017multi}} & \hfil{2.85} & \hfil{2.60} & \hfil{2.73} \\
\hfil{ARAN \cite{chen2019age}} &                    \hfil{2.77} & \hfil{\textbf{2.48}} & \hfil{2.63} \\
\hfil{BridgeNet \cite{li2019bridgenet}} &           \hfil{2.74} & \hfil{2.51} & \hfil{2.63} \\
\hfil{ADPF (ours)} &                                \hfil{\textbf{2.63}} & \hfil{2.50} & \hfil{\textbf{2.56}} \\
\bottomrule
\end{tabular}
\end{center}
\end{table}

\begin{table}
\ra{1.20}
\begin{center}
\caption{MAE values for several state-of-the-art Face-based Age Estimation Methods on the MORPH II Dataset under Setting III. Params indicates the number of parameters.}
\label{tab:morph3}
\begin{tabular}{p{0.18\textwidth}p{0.11\textwidth}p{0.11\textwidth}}\toprule
\hfil{Method} & \hfil{Params} & \hfil{MAE}\\ \midrule
\hfil{AGES \cite{geng2007automatic}} & \hfil{-} & \hfil{8.83} \\
\hfil{MTWGP \cite{zhang2010multi}} & \hfil{-} & \hfil{6.28} \\
\hfil{CA-SVR \cite{chen2013cumulative}} & \hfil{-} & \hfil{5.88} \\
\hfil{DLA \cite{wang2015deeply}} & \hfil{~ 6M} & \hfil{4.77} \\
\hfil{Rothe \textit{et al.} \cite{rothe2016some}} & \hfil{~ 20M} & \hfil{3.45} \\
\hfil{DLDLF \cite{shen2019deep}} & \hfil{~ 14M} & \hfil{2.94} \\
\hfil{DRF \cite{shen2019deep}} & \hfil{~ 14M} & \hfil{2.80} \\
\hfil{deep-JREAE \cite{tian2021facial}} & \hfil{~ 138M} & \hfil{2.77} \\
\hfil{BridgeNet \cite{li2019bridgenet}} & \hfil{~ 120M} & \hfil{\textbf{2.38}} \\
\hfil{ADPF (ours)} & \hfil{~ 14M} & \hfil{2.71} \\
\bottomrule
\end{tabular}
\end{center}
\end{table}

\textbf{Evaluation Metrics}. Results are reported based on two metrics, Mean Absolute Error (MAE) and Cumulative Score (CS). The MAE measures the average absolute difference between the predicted age and the ground truth:
\begin{equation}\label{mae}
    MAE = \frac{\sum_{z'}^{z}e_{z'}}{z},
\end{equation}
where $e_{z'}$ is the absolute error between the predicted age $\hat{u_{z'}}$ and the input label $u_{z'}$ for the $z'$-th sample, and $z$ is the total number of testing samples. The CS measures the percentage of images that are correctly classified in a certain age range as:
\begin{equation}\label{cs}
    CS(v) = -\frac{z_v}{z}\times{100\%},
\end{equation}
where $Z_v$ is the number of images whose predicted age $\hat{u_{z}}$ is in the range of $[u_{z}-v, u_{z}+v]$ and $v$ is the age margin.

\textbf{Implementation Details}. ADPF is implemented based on the open-source deep learning framework Pytorch \cite{paszke2017automatic} and trained with the SGD algorithm with a batch size of 32. We first train the AttentionNet for 200 epochs and then the FusionNet for another 200 epochs with the parameters of the AttentionNet fixed. The initial learning rate for both networks is set to 0.1 and drops by a factor of 0.1 after every 50 epochs. When training the AttentionNet, we empirically set $\lambda$ in Eq. \ref{attentionnet} to 0.01. Following our prior work, we use 5 patches when comparing with other state-of-the-art methods. All experiments are run on a single NVIDIA GTX 2080Ti GPU. To have a fair comparison against our prior work, we replace the age regression model used by our prior work with the age estimation loss in Eq. \ref{ae}.

\begin{figure}[t]
\begin{center}
\includegraphics[width=1.\linewidth]{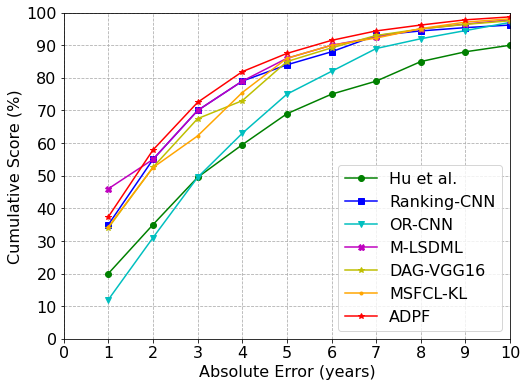}
\end{center}
   \caption{CS curves for several state-of-the-art Face-based Age Estimation Methods on the MORPH II Dataset under Setting I.}
\label{fig:morph1}
\end{figure}

\begin{figure}[t]
\begin{center}
\includegraphics[width=1.\linewidth]{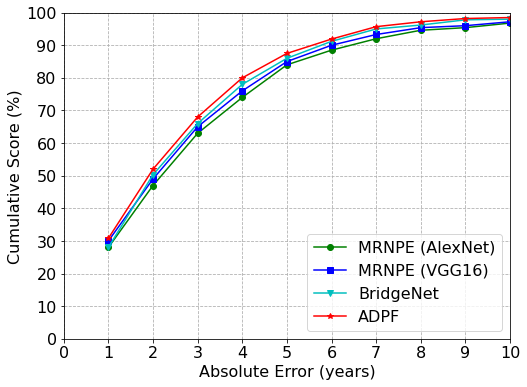}
\end{center}
   \caption{CS curves for several state-of-the-art Face-based Age Estimation Methods on the MORPH II Dataset under Setting II.}
\label{fig:morph2}
\end{figure}

\subsection{Evaluations on the MORPH II Dataset}

\begin{figure}[t]
\begin{center}
\includegraphics[width=1.\linewidth]{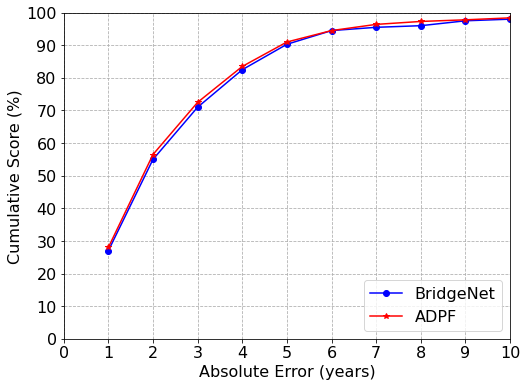}
\end{center}
   \caption{CS curves for several state-of-the-art Face-based Age Estimation Methods on the MORPH II Dataset under Setting III.}
\label{fig:morph3}
\end{figure}

The MAE values for the three aforementioned settings of the MORPH II dataset are tabulated in Table \ref{tab:morph1}-\ref{tab:morph3}, respectively. In Table \ref{tab:morph2}, the headings indicate the subsets used to compute the results. For example, \textit{S1/S2+S3} indicates the model is trained on the \textit{S1} subset and evaluated on the \textit{S2} and \textit{S3} subsets, and the \textit{Average} column tabulates the mean value of the two columns on the left. The CS curves for the three settings are presented in Fig. \ref{fig:morph1}-\ref{fig:morph3}, respectively. Note that not all methods report the results under this metric. As can be seen from these tables and figures, ADPF outperforms all state-of-the-art methods that focus on improving the feature extractor like the DAG family (DAG-GoolgeNet and DAG-VGG16) \cite{taheri2019use}, MSFCL family (MSFCL, MSFCL-LR, and MSFCL-KL) \cite{xia2020multi}, and our prior work \cite{wang2018fusion}. Also note that ADPF achieves comparable results to other methods that use customized estimators. For all three settings, the superior performance demonstrate that ADPF can predict ages accurately regardless of the imbalanced data distribution caused by other information like race. We also include comparisons of the number of parameters in Table II, IV, and V to provide more information into the performance of the evaluated methods. As tabulated, our method is among the ones with the smallest number of parameters, using only ~12\% of the total number of parameters used by the BridgeNet.

\begin{table}
\ra{1.20}
\begin{center}
\caption{MAE values for several state-of-the-art Face-based Age Estimation Methods on the FG-NET Dataset. Params indicates the number of parameters.}
\label{tab:fg}
\begin{tabular}{p{0.18\textwidth}p{0.11\textwidth}p{0.11\textwidth}}\toprule
\hfil{Method} & \hfil{Params} & \hfil{MAE}\\ \midrule
\hfil{AGES \cite{geng2007automatic}} & \hfil{-} & \hfil{6.77} \\
\hfil{IIS-LLD \cite{geng2013facial}} & \hfil{-} & \hfil{5.77} \\
\hfil{LARR \cite{guo2008image}} & \hfil{-} & \hfil{4.87} \\
\hfil{Feng \textit{et al.} \cite{feng2016human}} & \hfil{-} & \hfil{5.05} \\
\hfil{BIF \cite{guo2009human}} & \hfil{-} & \hfil{4.77} \\
\hfil{CPNN \cite{geng2013facial}} & \hfil{-} & \hfil{4.76} \\
\hfil{DEX \cite{rothe2018deep}} & \hfil{120M} & \hfil{4.63} \\
\hfil{CS-LBFL \cite{lu2015cost}} & \hfil{-} & \hfil{4.43} \\
\hfil{CS-LBMFL \cite{lu2015cost}} & \hfil{-} & \hfil{4.36} \\
\hfil{Mean-Variance Loss \cite{pan2018mean}} & \hfil{20M} & \hfil{4.10} \\
\hfil{GA-DFL \cite{liu2017group}} & \hfil{138M} & \hfil{3.93} \\
\hfil{LSDML \cite{liu2017label}} & \hfil{44M} & \hfil{3.92} \\
\hfil{ARAN \cite{chen2019age}} & \hfil{414M} & \hfil{3.79} \\
\hfil{M-LSDML \cite{liu2017label}} & \hfil{44M} & \hfil{3.74} \\
\hfil{DLDLF \cite{shen2019deep}} & \hfil{14M} & \hfil{3.71} \\
\hfil{DRF \cite{shen2019deep}} & \hfil{14M} & \hfil{3.47} \\
\hfil{DAG-VGG16 \cite{taheri2019use}} & \hfil{24M} & \hfil{3.08} \\
\hfil{DAG-GoogleNet \cite{taheri2019use}} & \hfil{131M} & \hfil{3.05} \\
\hfil{BridgeNet \cite{li2019bridgenet}} & \hfil{120M} & \hfil{\textbf{2.56}} \\
\hfil{ADPF (ours)} & \hfil{14M} & \hfil{2.86} \\
\bottomrule
\end{tabular}
\end{center}
\end{table}

\begin{table}
\ra{1.20}
\begin{center}
\caption{MAE values for several state-of-the-art Face-based Age Estimation Methods on the CACD.}
\label{tab:cacd}
\begin{tabular}{cp{0.08\textwidth}p{0.08\textwidth}}\toprule
\multirow{2}{*}{Method} & \multicolumn{2}{c}{MAE} \\
\cmidrule{2-3}
&\hfil{train} &\hfil{val} \\ \midrule
\hfil{DEX \cite{rothe2018deep}} &                    \hfil{4.79} & \hfil{6.52} \\
\hfil{DLDLF \cite{shen2019deep}} &                   \hfil{4.68} & \hfil{6.16} \\
\hfil{DRF \cite{shen2019deep}} &                     \hfil{\textbf{4.61}} & \hfil{5.63} \\
\hfil{ADPF (ours)} &                                 \hfil{4.72} & \hfil{\textbf{5.39}} \\
\bottomrule
\end{tabular}
\end{center}
\end{table}

\begin{figure}[t]
\begin{center}
\includegraphics[width=1.\linewidth]{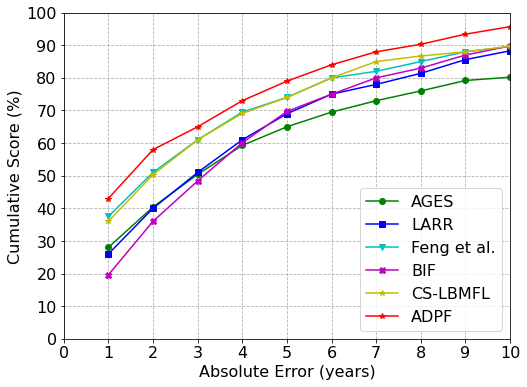}
\end{center}
   \caption{CS curves for several state-of-the-art Face-based Age Estimation Methods on the FG-NET Dataset.}
\label{fig:fg}
\end{figure}

\subsection{Evaluations on the FG-NET Dataset}

The MAE values and the CS curve are tabulated in Table \ref{tab:fg} and depicted in Fig. \ref{fig:fg}, respectively, for the FG-NET dataset. Again, not all methods report the results under the CS metric for the FG-NET dataset. It can be seen from Table \ref{tab:fg} that ADPF achieves an MAE value under 3.00, which shows that it can perform well even with small datasets.

\subsection{Evaluations on the CACD}

Evaluation results for the CACD under the MAE metric are tabulated in Table  \ref{tab:cacd}. ADPF achieves the best performance when trained on the validation dataset but only achieves the third best performance when trained on the training set. This may due to the age labels in the training set not being accurate. Since the input to the FusionNet of ADPF is sixfold, i.e., it includes one facial image and five patches, compared to other single-input networks, inaccurate labels may confuse the model due to mis-information. 

\subsection{Ablation Study}

% no ranking
% hybrid -> self

We conduct ablation experiments to demonstrate the effectiveness of each component of ADPF. Specifically, we aim to demonstrate that: 1) the hybrid attention mechanism is more effective than the self-attention mechanism when discovering age-specific patches; 2) the ranking operation in RMHHA is beneficial for feature learning in the FusionNet; 3) the effectiveness of the diversity loss; and 4) the importance of combining the FusionNet and the AttentionNet in a single framework. To this end, we design several baseline models as follows:

\begin{itemize}
  \item \textit{ADPF w/SA}: ADPF with the self-attention mechanism instead of the hybrid attention mechanism in the AttentionNet. The single channel feature maps are then generated by performing summation along the channel axis of the self-attention maps.
  \item \textit{ADPF w/o ranking}: ADPF without the ranking operation for age-specific patches.
  \item \textit{ADPF w/o diversity}: ADPF without the diversity loss.
  \item \textit{AttentionNet}: ADPF with no FusionNet.
\end{itemize}

\begin{figure}[t]
\begin{center}
\includegraphics[width=1.\linewidth]{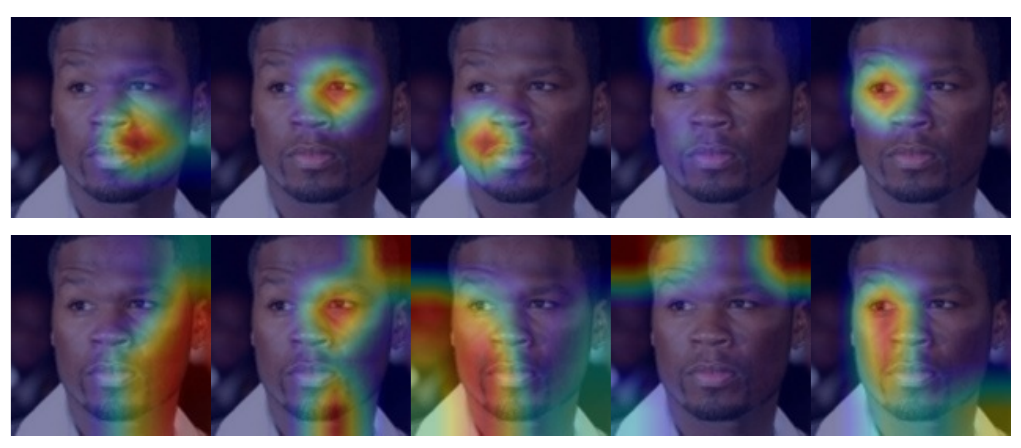}
\end{center}
   \caption{Attention maps computed by (upper row) the ADPF framework and (bottom row) the \textit{ADPF w/SA} baseline model.}
\label{fig:baseline}
\end{figure}

The evaluation results on the MORPH II dataset, Setting I, for the aforementioned baseline models and ADPF are tabulated in Table \ref{tab:baseline}. Example attention maps computed by the \textit{ADPF w/SA} baseline model are shown in Fig. \ref{fig:baseline}. As shown in this figure, although \textit{ADPF w/SA} can reveal key regions for age estimation, it may also reveal non-important regions, including sections of the background, which may be treated as noise during the feature learning process and eventually hinder the performance. In \textit{ADPF w/o ranking}, we feed the patches into the FusionNet based on their original order in the input tensor along the channel axis as produced by RMHHA. This feeding strategy cannot guarantee that the learning path for the most informative patch is long enough to extract meaningful features.

\begin{figure}[t]
\begin{center}
\includegraphics[width=1.\linewidth]{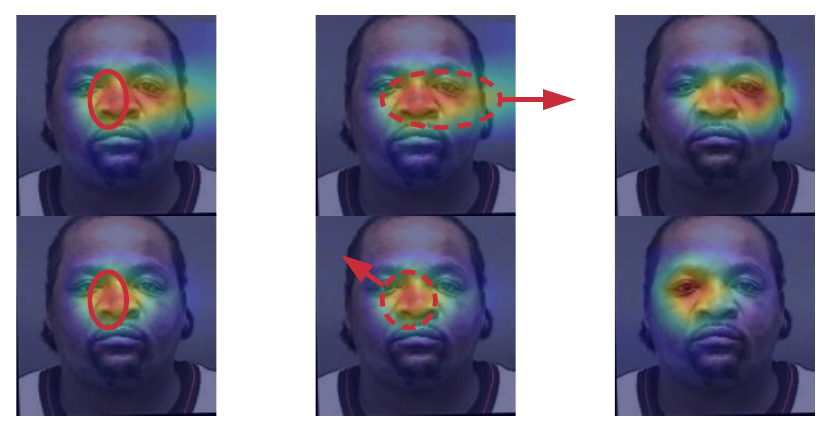}
\end{center}
   \caption{\textbf{Left}: Two attention maps overlap in the annotated area with out the supervision from the diversity loss. \textbf{Middle}: By minimizing the diversity loss, the two attention maps are forced to move in opposite directions. \textbf{Right}: attention maps generated by using the diversity loss.}
\label{fig:heat_map}
\end{figure}

To demonstrate the effectiveness of the proposed diversity loss, we visualize the attention maps learned on the MORPH II dataset, Setting I, by ADPF and the baseline model \textit{ADPF w/o diversity}. As shown in Fig. \ref{fig:heat_map}, in the \textit{ADPF w/o diversity} baseline model, the two attention maps overlap in the highlighted nose region, which leads to redundant input information to the network. With the aid of the diversity loss, these key regions detected by these two attention maps are forced to move in opposite directions resulting in two attention maps with negligible overlap. 

MAE values tabulated in Table \ref{tab:baseline} confirm the importance of combining the AttentionNet and the FusionNet in a single framework instead of using the AttentionNet exclusively. As we can see from this table, the performance of the \textit{AttentionNet} baseline model significantly drops compared to that of ADPF. This is mainly due to the limited number of feature maps available to the FC layer in the AttentionNet. With such a limited number of feature maps, the estimator cannot get enough information from the feature extractor. However, implementing the AttentionNet in this way is essential to learn and rank multiple single-channel attention maps, which shows the importance of combining the AttentionNet and the FusionNet in a single framework.

\begin{table}[t]
\ra{1.20}
\begin{center}
\caption{MAE values for several baseline models and the complete ADPF framework on the MORPH II Dataset under Setting I.}
\label{tab:baseline}
\begin{tabular}{p{0.18\textwidth}p{0.11\textwidth}}\toprule
\hfil{Method} & \hfil{MAE}\\ \midrule
\hfil{ADPF w/SA}     &                  \hfil{2.90} \\
\hfil{ADPF w/o ranking} &                \hfil{2.74} \\
\hfil{ADPF w/o diversity} &              \hfil{2.65} \\
\hfil{AttentionNet} &                    \hfil{3.31} \\
\hfil{ADPF} &                                     \hfil{\textbf{2.54}} \\
\bottomrule
\end{tabular}
\end{center}
\end{table}

% \begin{table}
% \ra{1.20}
% \begin{center}
% \caption{MAE Comparison between our prior work and the ADPF on the CACD.}
% \label{tab:patch_comparison}
% \begin{tabular}{cp{0.08\textwidth}p{0.08\textwidth}}\toprule
% \multirow{2}{*}{Method} & \multicolumn{2}{c}{MAE} \\
% \cmidrule{2-3}
% &\hfil{train} &\hfil{val} \\ \midrule
% \hfil{FusionNet \cite{wang2018fusion}} &             \hfil{4.91} & \hfil{5.76} \\
% \hfil{ADPF} &                                        \hfil{\textbf{4.73}} & \hfil{\textbf{5.38}} \\
% \bottomrule
% \end{tabular}
% \end{center}
% \end{table}

% \begin{table}[t]
% \ra{1.20}
% \begin{center}
% \caption{MAE Comparison between the \textit{w/o diversity} and the ADPF on the MORPH II Dataset under Setting I.}
% \label{tab:heat_map}
% \begin{tabular}{p{0.18\textwidth}p{0.11\textwidth}}\toprule
% \hfil{Method} & \hfil{MAE}\\ \midrule
% \hfil{\textit{w/o diversity}} &                   \hfil{2.65} \\
% \hfil{ADPF} &                                     \hfil{\textbf{2.55}} \\
% \bottomrule
% \end{tabular}
% \end{center}
% \end{table}

\subsection{Discussions}

\begin{table}
\ra{1.20}
\begin{center}
\caption{Training time of the FusionNet in our prior work and ADPF.}
\label{tab:training_efficency}
\begin{tabular}{p{0.18\textwidth}p{0.11\textwidth}p{0.11\textwidth}}\toprule
\hfil{Method} & \hfil{Hours} & \hfil{MAE}\\ \midrule
\hfil{BIF + FusionNet \cite{wang2018fusion}} &        \hfil{70}                &         \hfil{2.76} \\
\hfil{ADPF} &                                   \hfil{\textbf{25}}       &         \hfil{\textbf{2.54}} \\
\bottomrule
\end{tabular}
\end{center}
\end{table}

\begin{figure}[t]
\begin{center}
\includegraphics[width=1.\linewidth]{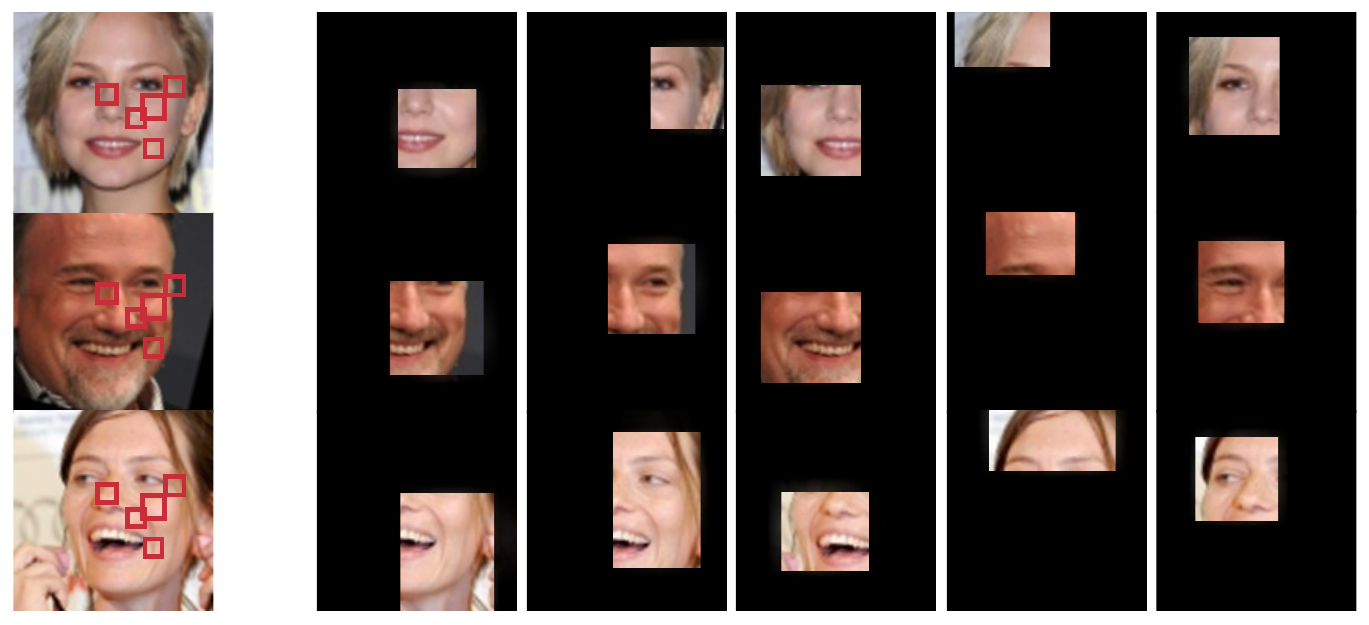}
\end{center}
   \caption{Sample age-specific patches computed by our prior work \cite{wang2018fusion} and the ADPF framework. The left column depicts the original facial images with patches computed by \cite{wang2018fusion} highlighted in red. The five patches computed by the ADPF framework are depicted in the last five columns. Within these columns, the patches are depicted from left to right in descending order in terms of their importance.}
\label{fig:patch_comparison}
\end{figure}

% number of heads (discussion)

\subsubsection{Training Efficiency}
We compare the training time required by our prior work \cite{wang2018fusion} and the ADPF on the MORPH II dataset with \textit{Setting I}. The training times are tabulated in Table \ref{tab:training_efficency}. Note that it takes about 70 hours to train the whole method in \cite{wang2018fusion} out of which 60 hours are required to compute and rank BIF-based patches and 10 hours to train the CNN. Thanks to the proposed RMHHA mechanism, ADPF only takes about one third of this time to converge with significantly boosted performance (see MAE values). In addition, the process of acquiring patches and training the CNN can only be done separately in \cite{wang2018fusion}. On the contrary, in ADPF, the training of the FusionNet can be done directly after the AttentionNet converges, which further makes the training process more time-efficient.

\subsubsection{Robustness of Age-Specific Patches}
We visually compare the patches computed by the BIF and Adaboost algorithms used in \cite{wang2018fusion} and those computed by RMHHA. This comparison is conducted on the CACD dataset as the facial images in this dataset contain PIE variations. Fig. \ref{fig:patch_comparison} depicts sample patches, where the most informative patches computed by \cite{wang2018fusion} are marked with red boxes. It is clear that the location and shape of each patch computed by \cite{wang2018fusion} are identical for all the images. On the contrary, the location and shape of the patches computed by the RMHHA vary from image to image. For example, in the bottom row, the patch capturing the right laughline is larger than that of the other two images, which allows capturing the complete skin texture of this key region. 

\subsubsection{Number of Heads}
The performance of ADPF with different number of attention heads is tabulated in Table \ref{tab:heads}. We can see that the best performance can be achieved when 5 or 6 attention heads are implemented. This may due to the fact that with less heads, some age-specific patches may remain undiscovered. Moreover, since most of the facial regions are already revealed when 5 attention heads are used, adding more heads only forces the framework to attend to irrelevant regions like the background, which as discussed previously, can be treated as noise and degrade the performance. Since 6 heads requires more time to train with no significant performance gains, 5 is an appropriate number to be used by ADPF. 

\begin{table}
\ra{1.20}
\begin{center}
\caption{Performance of ADPF with different number of attention heads on the MORPH II dataset under Setting I.}
\label{tab:heads}
\begin{tabular}{p{0.10\textwidth}p{0.03\textwidth}p{0.03\textwidth}p{0.03\textwidth}p{0.03\textwidth}p{0.03\textwidth}p{0.03\textwidth}}\toprule
\hfil{\# Heads} & \hfil{3} & \hfil{4}& \hfil{5} & \hfil{6}& \hfil{7} & \hfil{8}\\ \midrule
\hfil{MAE} & \hfil{2.77} & \hfil{2.62} & \hfil{\textbf{2.54}} & \hfil{\textbf{2.54}} & \hfil{2.55} & \hfil{2.61}\\    
\bottomrule
\end{tabular}
\end{center}
\end{table}

% \subsubsection{\textcolor{blue}{Limitations}}
% \textcolor{blue}{Although the training efficiency has dramatically increased compared to our prior work \cite{wang2018fusion}, the pipeline requires more time during inference since the involvement of the hybrid attention mechanism. Such attention mechanisms can have a quadratic complexity with respect to the input dimension due to the similarity comparison operations \cite{babiloni2021poly}.}

\section{Conclusion}

In this paper, we proposed the ADPF framework to improve the performance of the face-based age estimation task. Our framework merges an AttentionNet and a FusionNet. The AttentionNet includes a novel hybrid attention mechanism, namely  RMHHA, which allows learning multiple single-channel attention maps to reveal age-specific patches. After ranking them, these patches are used by the FusionNet, along with the facial image to compute the final age prediction. Based on evaluations on several benchmark datasets, ADPF significantly improves prediction accuracy compared to several state-of-the-art methods. ADPF also outperforms our previous work, both in terms of accuracy and training times. Since this work focuses on building customized feature extractors, in the future, we will investigate the design of customized estimators to further boost performance by, for example, considering the ordinal information among ages and further minimizing the distance between label distributions and feature distributions.

\section{Acknowledgment}

This work was supported by the EU Horizon 2020 - Marie Sklodowska-Curie Actions through the project Computer Vision
Enabled Multimedia Forensics and People Identification (Project No. 690907, Acronym: IDENTITY).

\ifCLASSOPTIONcaptionsoff
  \newpage
\fi



%\bibliographystyle{IEEEtran}

%-------------------------------------------------------------------------
%%%%%%%%% References
\bibliographystyle{IEEEbib}
\bibliography{refs}

\end{document}